\documentclass[
]{ceurart}

\usepackage{subcaption}

\begin{document}

\copyrightyear{2020}
\copyrightclause{Copyright for this paper by its authors.
  Use permitted under Creative Commons License Attribution 4.0
  International (CC BY 4.0).}
  
\conference{IberLEF'20: Iberian Languages Evaluation Forum,
  September 2020, Málaga, Spain}
  
\title{NLNDE at CANTEMIST: Neural Sequence Labeling and Parsing Approaches for Clinical Concept Extraction}

\author[1,2]{Lukas Lange}[%
email=lukas.lange@de.bosch.com,
]
\address[1]{Bosch Center for Artificial Intelligence,
  Robert-Bosch-Campus 1, Renningen, 71272, Germany}
\address[2]{Saarland University, 
  Saarland Informatics Campus, Saarbrücken, 66123, Germany}

\author[1,3]{Xiang Dai}[%
email=dai.xiang.au@gmail.com,
]
\address[3]{University of Sydney, Sydney, 2006, Australia}

\author[1]{Heike Adel}[%
email=heike.adel@de.bosch.com,
]

\author[1]{Jannik Strötgen}[%
email=jannik.stroetgen@de.bosch.com,
]

\newcommand{\fscore}[1][1]{$F_{#1}$\xspace}
\begin{abstract}
  The recognition and normalization of clinical information, such as tumor morphology mentions, is an important, but complex process consisting of multiple subtasks. 
  In this paper, we describe our system for the CANTEMIST shared task, which is able to extract, normalize and rank ICD codes from Spanish electronic health records 
  using neural sequence labeling and parsing approaches with context-aware embeddings. Our best system achieves 85.3 \fscore, 76.7 \fscore, and 77.0 MAP for the three tasks, respectively.
\end{abstract}


\maketitle

\section{Introduction}

Collecting and understanding key clinical information, such as disorders, symptoms, drugs, etc., from electronic health records (EHRs) has wide-ranging applications within clinical practice and transnational research~\cite{Leaman:Khare:JBI:2015,Wang:Wang:JBI:2018}.
A better understanding of this information can facilitate novel clinical studies on the one hand, and help practitioners to optimize clinical workflows on the other hand.
For example, to improve clinical decision support and personalized care of cancer patients, \citet{Jensen:Ruiz:SR:2017} developed a methodology to estimate disease trajectories from EHRs, which can predict 80\% of patient events in advance.
However, free text is ubiquitous in EHRs. This leads to great difficulties in harvesting knowledge from EHRs.
Therefore, natural language processing (NLP) systems, especially information extraction components, play a critical role in extracting and encoding information of interest from clinical narratives, as this information can then be fed into downstream applications.

\begin{figure}
  \centering
  \includegraphics[width=0.9\linewidth]{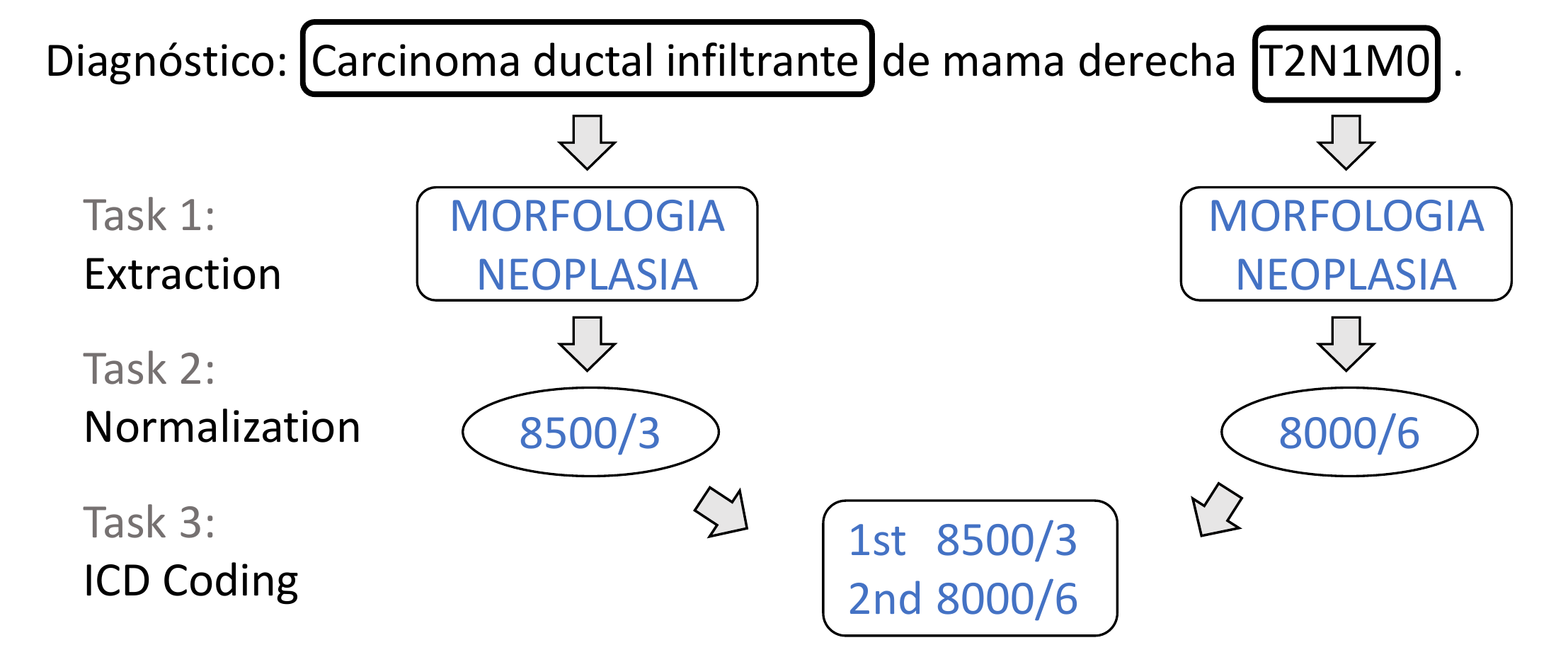}
  \caption{Sample sentence with normalized and ranked extractions.}
  \label{fig:example}
\end{figure}

The CANcer TExt Mining Shared Task (CANTEMIST) \cite{CantemistOverview} focuses on identifying a critical type of concept related to cancer, namely tumor morphology. There are three independent subtasks as shown in Figure~\ref{fig:example}: (1) The extraction of tumor mentions, (2) the subsequent normalization to ICD codes and (3) the ranking by importance of the codes for each document.

In this paper, we describe our submission as Neither Language Nor Domain Experts (NLNDE) to the shared task.
We treat the first subtask as a named entity recognition (NER) task and use neural sequence labeling and parsing approaches as frequently done to address NER in low resource settings. 
For the other two subtasks, we use rather simple non-deep learning methods, due to the very limited amount of training data: For the second subtask, the extracted entities are normalized using string matching and Levenshtein distance and the ranking of the third subtask is based on frequency.

\section{Related Work}

To identify medical concepts within the clinical narratives in EHRs, several machine learning-based named entity recognition (NER) and normalization systems were implemented~\cite{Leaman:Khare:JBI:2015,Leaman:Dogan:BioInfo:2013,Leaman:Lu:BioInfo:2016}.
Current state-of-the-art models for the extraction of clinical concepts are typically implemented as recurrent neural networks based on multiple different embeddings \cite{gonzalez-agirre-etal-2019-pharmaconer,lange-etal-2020-closing}. 
DNorm, introduced in~\cite{Leaman:Dogan:BioInfo:2013}, applied a pairwise learning to rank approach to automatically learn a mapping from disease mentions to disease concepts from the training data. Evaluation results show that the machine learning method can effectively model term variations and achieves much better results than traditional techniques based on lexical normalization and matching, such as MetaMap~\cite{Aronson:AMIA:2001}. \citet{Leaman:Khare:JBI:2015} introduced an extension of DNorm, called DNorm-C, which approaches both discontinuous NER and normalization using a pipeline approach. 
A joint model for NER and normalization was introduced in~\cite{Leaman:Lu:BioInfo:2016}, aiming to overcome the cascading errors caused by the pipeline approach and enable the NER component to exploit the lexical information provided by the normalization component.

Other efforts on addressing both medical NER and normalization in other text types also exist. \citet{Metke-Jimenez:Karimi:BMDID:2016} compared different techniques for identifying medical concepts and drugs from medical forums. 
\citet{Zhao:Liu:AAAI:2019} proposed a deep neural multi-task learning method to jointly model NER and normalization from biomedical publications, where stacked recurrent layers are shared among different tasks, enabling mutual support between tasks. 
Similarly, \citet{Lou:Zhang:BioInfo:2017} proposed a transition-based model to jointly perform disease NER and normalization, combined with beam search and online structured learning. 
Experiments show that their joint model performs well on PubMed abstracts.

In contrast to concept normalization, which identifies a one-to-one mapping between text snippet and medical concept, ICD coding assigns most relevant ICD codes to a document as a whole~\cite{Pestian:Brew:BioNLP:2007,Neveol:Robert:CLEF:2018}. 
Most previous methods simplified this task as a text classification problem, and built classifiers using CNNs~\cite{Karimi:Dai:BioNLP:2017} or tree-of-sequences LSTMs~\cite{Xie:Xing:ACL:2018}. 
Since ICD codes are organized under a hierarchical structure, \citet{Mullenbach:Wiegreffe:NAACL:2018} and \citet{Cao:Chen:ACL:2020} proposed models to exploit code co-occurrence using label attention mechanism and graph convolutional networks, respectively.
\section{Approach}
This section provides an overview of the different methods tested for the three tasks, starting with the extraction, followed by the normalization and finally the ranking of the entities. Our architecture for the complete sequence of all three tasks is shown in Figure~\ref{fig:system}.

\newcommand{\subA}{\textit{S1}\xspace}
\newcommand{\subB}{\textit{S2}\xspace}
\newcommand{\subC}{\textit{S3}\xspace}
\newcommand{\subD}{\textit{S4}\xspace}
\newcommand{\subE}{\textit{S5}\xspace}

\begin{figure}
  \centering
  \includegraphics[width=\linewidth]{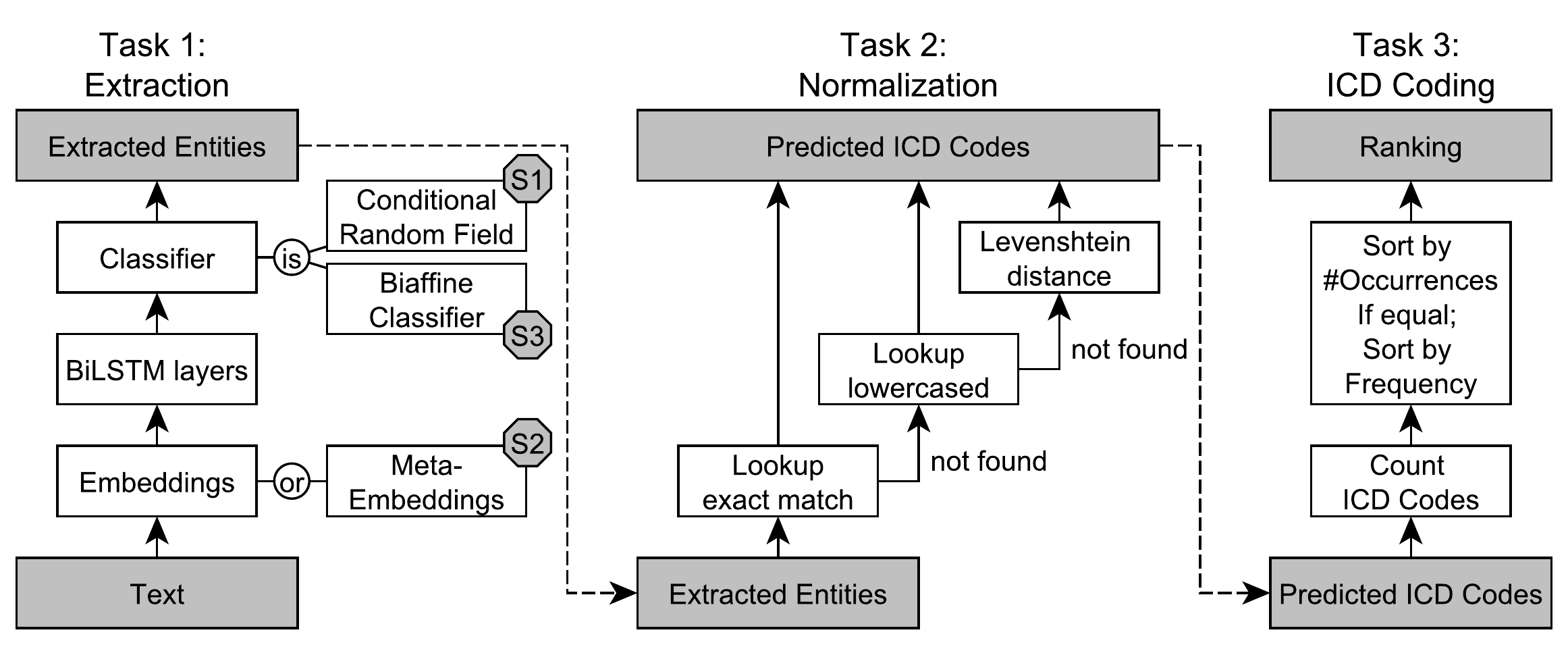}
  \caption{Overview of the NLNDE system architecture. \subA, \subB and \subC are system variants in our different submissions. 
}
  \label{fig:system}
\end{figure}

\subsection{Task 1: Named Entity Recognition}
We mainly experiment with two different methods for the extraction of tumor mentions. The first model treats the extraction as a sequence labeling problem without nested mentions, while the second model treats the problem as a parsing problem that allows the detection of nested mentions.

\paragraph{Sequence Labeling Model.}
For the sequence labeling model, the data is converted into the BIO format~\cite{ramshaw-marcus-1999-text} using SpaCy\footnote{\url{https://spacy.io/api/tokenizer}} as the tokenizer. Overlapping annotations are resolved to a single annotation by selecting the longest sequence. 
We use a recurrent neural network, in particular, a bidirectional long short-term memory network (BiLSTM) with a conditional random field (CRF) output layer similar to \cite{lample-etal-2016-neural}.
For our choice of embeddings, we follow \cite{lange-etal-2019-nlnde-meddocan} who used a similar system for de-identification of Spanish clinical documents.
In particular, we use pre-trained fastText embeddings \cite{bojanowski-etal-2017-enriching} that were trained on articles from Wikipedia and the Common Crawl, 
as well as domain-specific fastText embeddings \cite{soares-etal-2019-medical} that were pre-trained on articles of the Spanish online archive SciELO\footnote{\url{https://scielo.org/}} for clinical documents.
In addition, we include byte-pair-encoding embeddings \cite{heinzerling-strube-2018-bpemb} with 300 dimensions and a vocabulary size of 200,000 syllables.
Finally, we add pre-trained FLAIR embeddings \cite{akbik-etal-2018-contextual}, which are calculated by contextualized character language models.
All the $n$ different embeddings are concatenated into a single embedding vector $e$
\begin{align}
e_{CONCAT}(i) = [e_{1}(i); \cdots ; e_{n}(i) ]
\end{align}

The embeddings are then fed into a stacked BiLSTM network that generates the feature presentation $f$ given the embeddings $e$ for each word in the sentence. $f$ is then mapped to the size of the label space and fed into a conditional random field (CRF) classifier \cite{lafferty-etal-2001-crf} that computes the most probable sequence of labels. 
We found that 3 stacked LSTM layers with a hidden size of 128 units each worked best in our experiments. The stacking of up to three layers increased the extraction performance by more than 1 \fscore point compared to a single LSTM layer. 

\paragraph{Tokenization.}
We further analyze the effects of tokenization errors on the extraction. 
The BiLSTM-CRF using the SpaCy tokenizer achieves an \fscore of 82.4 on the development set (Precision (P): 84.9, Recall (R): 80.1).
We then derive the following custom splitting rules according to annotation boundary problems from the training data.
\begin{itemize}
    \item Suffix Rule: Cut off the suffix if the word is ending with a ``.'' or ``-''
    \item Prefix Rule: Cut off the prefix if the word starts with a ``-''
    \item Infix Rule: split each word at hyphens, punctuation and quotation marks into three parts.
\end{itemize}
The rules increase performance for all three metrics by 0.4--0.5 points (P: 85.4, R: 80.5, \fscore: 82.9).

\paragraph{Meta-Embeddings.}
Related work has shown significant improvements when the simple concatenation of embeddings is replaced with a different meta-embedding method.
We experiment with an attention mechanism as described by \citet{kiela-etal-2018-dynamic} to create meta-embeddings of several different embedding types.
Such meta-embeddings were shown to be useful in multiple extraction tasks \cite{kiela-etal-2018-dynamic, lange-etal-2020-choice, winata-etal-2019-hierarchical,lange-etal-2019-nlnde-pharmaconer}. 
As all embeddings have a different size of up to 2048 dimensions, all embeddings are mapped to the same space with dimension $E$ first. We set $E$ to the size of the largest embeddings. For this, we use a non-linear mapping $Q_i$ with bias $b_i$ for embedding $e_i$:
\begin{align}
x_i = \tanh(Q_i \cdot e_i + b_i)
\end{align}

We take the attention method proposed by \citet{lange-etal-2019-nlnde-pharmaconer} who used feature-based attention.
With this, the attention function has access to additional word information, in our case the word's shape, frequency and length. This helps to infer linguistic information about the word that can be useful for the attention weight computation but is not encoded in the word vectors. The features are added as a vector $f_w$ to the attention function:
\begin{align}
\alpha_i &= \frac{\exp(V \cdot \tanh(W x_i + f_w))}{\sum_{l=1}^n \exp(V \cdot \tanh(W x_l + f_w))} \\
e_{META} &= \sum_i \alpha_i \cdot x_i
\end{align}
with $x_i$ being the mapped embeddings $e_i$ and $V$ and $W$ being parameters of the model that are learnt during training.
The final meta-embedding $e_{META}$ is then used as input to the stacked BiLSTM network.  The meta-embedding model has a hidden size of 25 dimensions for the attention computation. 

\paragraph{Biaffine Classifier.}
Recently, a trend emerged of modeling different natural language processing tasks as parsing tasks and thus, solve them by using a dependency parser. Examples are named entity recognition \cite{yu-etal-2020-named} and negation resolution \cite{kurtz-etal-2020-end}.

We experiment with such a system and model the extraction task as a parsing task. For this, we replace the CRF classifier with a biaffine classifier \cite{dozat-manning-2017-deep}. 
Following \citet{yu-etal-2020-named}, we apply two separate feed-forward networks (FFNN) to the features $f$ generated from the stacked BiLSTM to create start and end representations of all possible spans ($h_s$/$h_e$).
Then, we use biaffine attention \cite{dozat-manning-2017-deep} over the sentence to compute the scores $r_m$ for each span $i$ in the sentence that could refer to a named entitiy.
\begin{align}
h_s(i) &= FFNNs(f_{s_i}) \\
h_e(i) &= FFNNe(f_{e_i}) \\
r_m(i) &= h_s^\top (i) U_m h_e(i) + W_m (h_s(i) \oplus h_e(i)) + b_m
\end{align}

Similar to \citet{yu-etal-2020-named}, we use multilingual BERT, character and fastText embeddings. We experimented with the same set of embeddings that we used for the BiLSTM-CRF model as well, but the performance decreased for the biaffine model.
Again, the embeddings are fed into the BiLSTM to obtain the word representations $f$. 
We found that 5 stacked LSTM layers with a size of 200 hidden units each worked best for the biaffine model. Using this combination of hyperparameters improved the model by roughly 1 \fscore point compared to the originally proposed model consisting of 3 layers of size 200.

For the BiLSTM-CRF and biaffine models we mostly follow the hyperparameter configurations and training routines of \citet{akbik-etal-2018-contextual} and \citet{yu-etal-2020-named}, respectively, with exceptions regarding the number and sizes of the recurrent layers mentioned above.

\subsection{Task 2: Normalization}
The second task requires the normalization of the previously extracted entities to ICD-O-3 codes (Spanish version: eCIE-O-3.1).
As a large number of possible ICD codes appears only once or never in the training data, we decided against deep-learning methods, as simply not enough training instances are available for this large label set. 
Instead, we use an approach based on string matching and Levenshtein distance \cite{levenshtein-1966-binary}.

For this, we collect all entities from the training set and their ICD code. As there is only little ambiguity among these entities, we use a context-independent method for the normalization. 
Using the entities from the training set, we are able to correctly assign 70\% of the ICD codes to entities from the development set using exact string matching with a very low false-positive rate ($< 1\%$). Using lower-cased matching, the number of correctly assigned codes slightly increases. 
Given that these methods assign codes almost perfectly to known entities, we first apply exact string matching and then lower-cased matching. 
For the remaining unmatched entities, we compute the Levenshtein distance between the given string and strings from the training data to find the closest neighbor among the known training instances and assign the corresponding code.
This method achieves 87\% \fscore on the gold extractions of the unseen development set.

\begin{table}
\centering
\caption{Results of different normalization methods on the development set.}
\label{tab:normalization}
\begin{tabular}{l|rrr|c} \toprule
Method & Correct & False & Unmatched & \fscore on gold extractions \\ \midrule
String matching         & 2341 &  21 & 979 & 70.07 \\
+ lowercased            & 2374 &  22 & 945 & 71.06 \\
++ Levenshtein distance & 2910 & 431 &   0 & 87.10 \\
\bottomrule
\end{tabular}
\end{table}

\subsection{Task 3: ICD Coding}
The purpose of the last subtask is the creation of a ranked list of ICD codes for a given document.
For this ICD coding, we create a ranking with a sorting function based on code frequency.
We sort by the number of times each code occurs in the given document under the assumption that codes that appear more often inside a document are more important.
Whenever two codes appeared an equal amount of times, they are ranked by their general frequency as found on the training set.
This method achieves a MAP of 73.82 using the gold extractions of the unseen development set.

\subsection{Submissions}
The following five runs are the NLNDE submissions to the CANTEMIST shared task. The difference between the runs lies in the model architecture used in the extraction track. The normalization and ICD coding methods are equal across the submissions and solely based on the predicted extraction of the first subtask:

\begin{itemize}
    \item [\subA]:  A BiLSTM-CRF model with a concatenation of FLAIR, fastText, BPEmb and domain-specific fastText embeddings.
    \item [\subB]:  A BiLSTM-CRF model with feature-based meta-embeddings as a replacement for the concatenation of embeddings used in \subA. 
    \item [\subC]:  A biaffine model with multilingual BERT and fastText embeddings for nested named entity recognition.
    \item [\subD]: A similar biaffine model trained on the development set in addition to the training set.
    \item [\subE]: An ensemble of \textit{S1/S2/S3} based on majority voting. Predictions are accepted into the ensemble classifier whenever at least two models predicted identical entity offsets. 
\end{itemize}
\section{Results}
The official results for the three tracks of the CANTEMIST shared task are shown in Tables~\ref{tab:res-ner}, \ref{tab:res-norm} and \ref{tab:res-coding}, respectively. The official evaluation metric of the test set is highlighted in gray and the best model is highlighted in bold.

\subsection{Results for Task 1 and 2: Named Entity Recognition and Normalization}
The \textbf{BiLSTM-CRF} (\subA) is a competitive baseline model for our experiments with 82.7 \fscore for the extraction and 72.9 \fscore for the normalization.
Even though the \textbf{meta-embeddings} (\subB) improve performance on the development set, at least for the normalization, we observe contrary results on the test data, as the concatenation of embeddings works better for this. 

The \textbf{biaffine model} (\subC) achieves a much higher precision than the BiLSTM-CRF with +2 \fscore points for the extraction and +1 \fscore point for the normalization on the development set. This gap further increases on the unseen test data. The difference in recall is not that large, even though the biaffine model is able to extract nested entities. However, the number of nested mentions is rather low and the ability to extract them does not seem to make a big difference in practice for this shared task. Overall, the biaffine model dominates because of the better precision, which might be explained by the fact that many of the tumor mentions cover multiple tokens and the parsing model is better in capturing those long-distant dependencies.
A more detailed analysis on this is provided in Section~\ref{sec:analysis}. 
In addition, the biaffine model can be further improved by training on a combination of training and development set, resulting in our best submission (\subD). 

The \textbf{ensemble model} (\subE) effectively increases the precision compared to the single models, in particular for the normalization, but it does not have the same recall, as only entities predicted by at least two of the three models get accepted into the output. Thus, only high-confidence entities are output by the ensemble classifier. As a result, this model may be the better choice if precision is preferred over recall.

\definecolor{Gray}{gray}{0.85}
\newcolumntype{a}{>{\columncolor{Gray}}c}
\newcommand{\best}[1]{\textbf{\textit{#1}}}

\begin{table}
\centering
\caption{Results of task 1: Extraction of tumor morphology mentions.}
\label{tab:res-ner}
\begin{tabular}{ll|ccc|cca} \toprule
 & & \multicolumn{3}{c|}{Dev} & \multicolumn{3}{c}{Test} \\
 & & P & R & \fscore & P & R & \fscore \\  \midrule
\subA & BiLSTM-CRF     & 84.7 & 82.4 & 83.6 & 82.4 & 83.0 & 82.7 \\
\subB & MetaEmbeddings & 84.7 & 82.4 & 83.6 & 81.5 & 82.3 & 81.9 \\
\subC & Biaffine       & 86.8 & 82.1 & 84.4 & 85.0 & 83.5 & 84.2 \\
\subD & Biaffine-Dev   &    - &    - &    - & \best{85.4} & \best{85.2} & \best{85.3} \\
\subE & Ensemble       & 87.8 & 81.3 & 84.4 & 84.7 & 80.8 & 82.7 \\ \bottomrule
\end{tabular}
\end{table}

\begin{table}
\centering
\caption{Results of task 2: Normalization of extractions to corresponding ICD-O-3 codes. }
\label{tab:res-norm}
\begin{tabular}{ll|ccc|cca|ccc} \toprule 
 & & \multicolumn{3}{c|}{Dev} & \multicolumn{3}{c|}{Test} & \multicolumn{3}{c}{Test w/o code 8000/6} \\
 & & P & R & \fscore & P & R & \fscore & P & R & \fscore \\ \midrule
\subA & BiLSTM-CRF     & 76.4 & 74.3 & 75.3 & 74.3 & 74.9 & 74.6 & 75.0 & 70.9 & 72.9 \\
\subB & MetaEmbeddings & 76.6 & 74.5 & 75.6 & 73.5 & 74.1 & 73.8 & 74.6 & 70.9 & 72.7 \\
\subC & Biaffine       & 79.0 & 74.7 & 76.8 & \best{76.7} & 75.3 & 76.0 & 76.4 & 71.4 & 73.8 \\
\subD & Biaffine-Dev   &    - &    - &    - & \best{76.7} & \best{76.6} & \best{76.7} & 77.3 & \best{72.6} & \best{74.9} \\
\subE & Ensemble       & 80.0 & 74.0 & 76.9 & \best{76.7} & 73.2 & 74.9 & \best{77.4} & 70.2 & 73.6 \\ \bottomrule
\end{tabular}
\end{table}

\begin{table}
\centering
\caption{Results of task 3: Creating a ranked coding of the given document. }
\label{tab:res-coding}
\begin{tabular}{ll|c|ccca|cccc} \toprule
 & & \multicolumn{1}{c|}{Dev} & \multicolumn{4}{c|}{Test} & \multicolumn{4}{c}{Test w/o code 8000/6} \\
 & & MAP & P & R & \fscore & MAP & P & R & \fscore & MAP \\ \midrule
\subA & BiLSTM-CRF     & 74.2 & 75.5 & 76.2 & 75.9 & 73.7 & 72.7 & 72.1 & 72.4 & 69.7 \\
\subB & MetaEmbeddings & 74.4 & 74.8 & 75.8 & 75.3 & 73.5 & 71.9 & 71.6 & 71.8 & 69.4 \\
\subC & Biaffine       & 75.0 & 75.9 & 76.3 & 76.1 & 73.9 & 73.0 & 72.2 & 72.6 & 70.2 \\
\subD & Biaffine-Dev   & -    & 77.0 & \best{77.1} & \best{77.0} & \best{74.9} & 74.3 & \best{72.8} & \best{73.6} & \best{71.4} \\
\subE & Ensemble       & 74.2 & \best{77.2} & 74.9 & 76.0 & 73.1 & \best{74.6} & 70.7 & 72.6 & 69.3 \\ \bottomrule
\end{tabular}
\end{table}

\subsection{Results for Task 3: ICD Coding}
The results for the third subtask, the ranked coding, are close to the results on the gold extractions. This indicates that the systems are able to extract the most important entities correctly. 
Overall, the differences between the systems are rather small as shown in Table~\ref{tab:res-coding}.
For example, the MAP score for the biaffine model (\subC) is only 0.2 points higher than the BiLSTM-CRF (\subA). Only the biaffine model trained on the combination of training and development data (\subD) achieves a slightly higher performance of up to a MAP score of 77.0.

Following the official evaluation, we include the results without the most frequent code "8000/6" (Metastatic Cancer) for the normalization and coding tasks in Tables \ref{tab:res-norm} and \ref{tab:res-coding}. With this, we observe a performance drop for all submissions between 2 and 3 \fscore or MAP points.

To conclude, our results show that the individual task-specific components deliver good results on the development as well as on the test set. Furthermore, the sequential execution as a pipeline model of extraction, normalization and ranking works well in practice.

\subsection{Analysis: BiLSTM-CRF vs. Biaffine Classifier}\label{sec:analysis}
\begin{figure}
     \centering
     \begin{subfigure}[b]{0.3\textwidth}
         \centering
         \includegraphics[width=\textwidth]{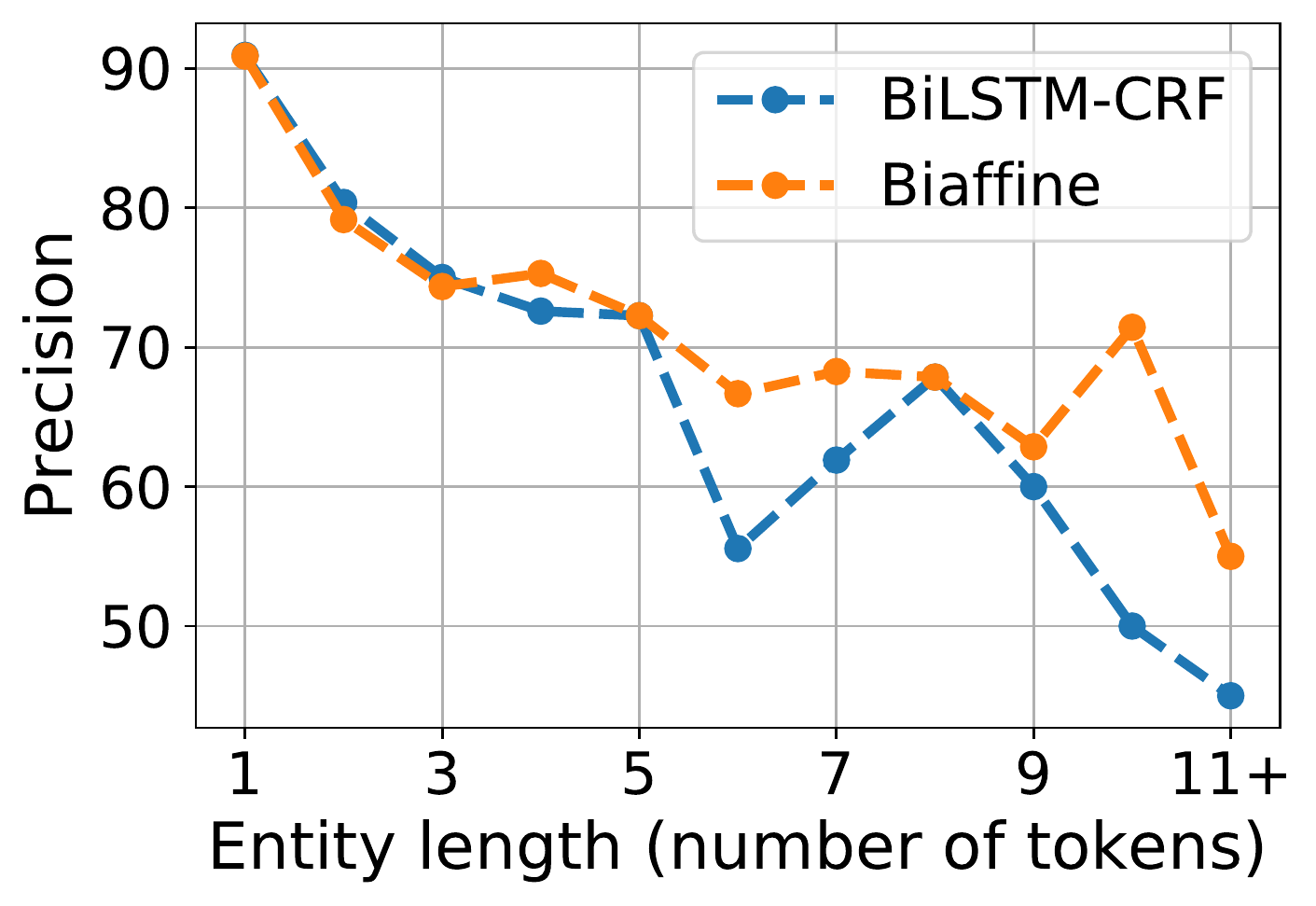}
         \caption{Task 1: Extraction}
         \label{fig:length-ner}
     \end{subfigure}
     \hfill
     \begin{subfigure}[b]{0.3\textwidth}
         \centering
         \includegraphics[width=\textwidth]{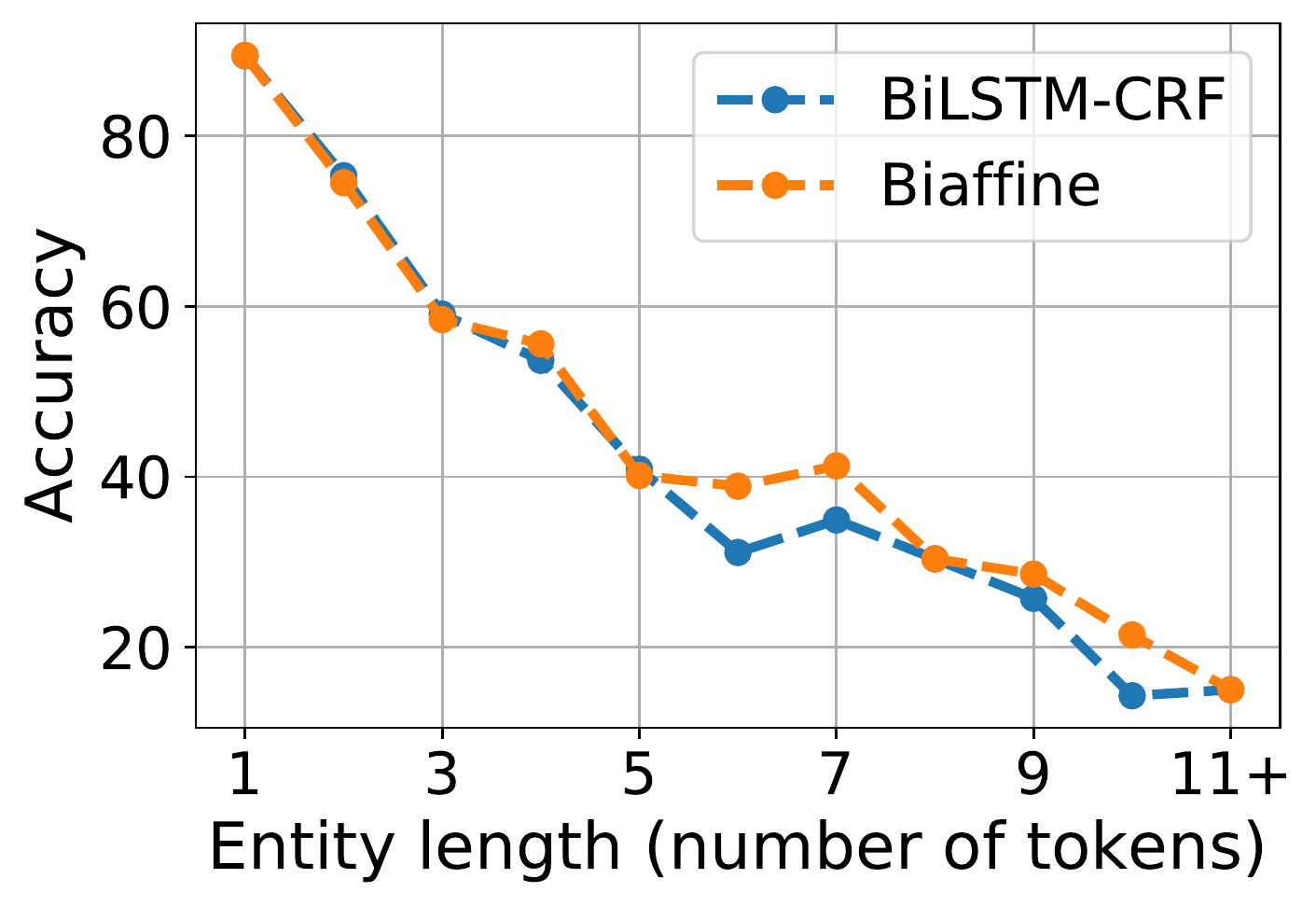}
         \caption{Task 2: Normalization}
         \label{fig:length-norm}
     \end{subfigure}
     \hfill
     \begin{subfigure}[b]{0.3\textwidth}
         \centering
         \includegraphics[width=\textwidth]{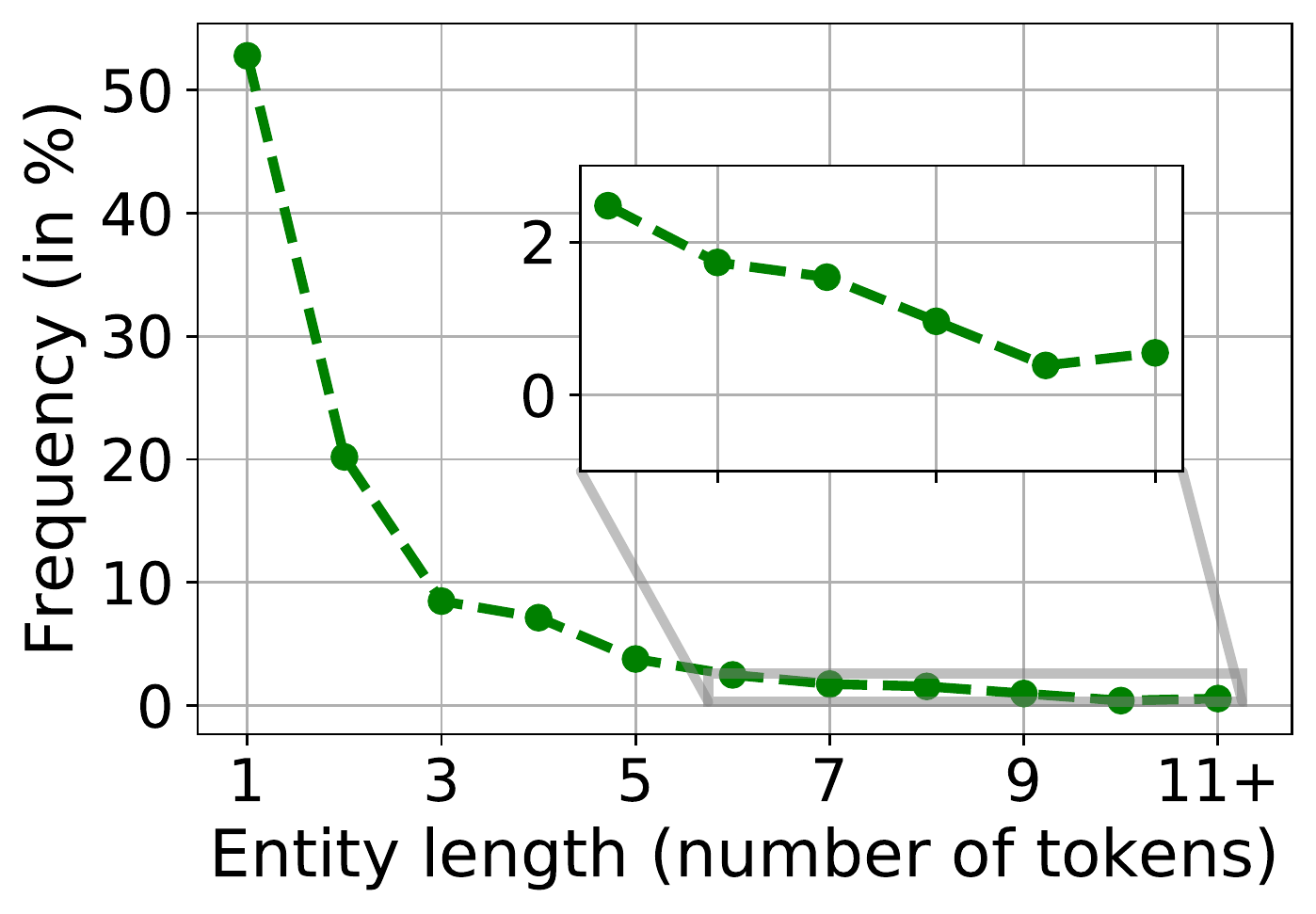}
         \caption{Relative frequency}
         \label{fig:length-freq}
     \end{subfigure}
    \caption{Results for entities of different lengths. (a) displays the impact of entity length on the extraction and (b) for the normalization. In (c) the relative frequency of these entities is shown. The last data point (11+) in all plots is the aggregation of all entities longer than 10 tokens.}
    \label{fig:length-analysis}
\end{figure}

In the following, the performance differences between the BiLSTM-CRF and biaffine models are analyzed with a focus on the lengths of the entities. As shown in Table~\ref{tab:res-ner}, the main difference lies in the higher precision of the biaffine model. Figure~\ref{fig:length-ner} shows the precision for entities with respect to their length. In particular, for shorter entities, there are no differences in performance between the two model architectures. Starting with entities consisting of 6 and more tokens, the biaffine model begins to outperform the BiLSTM-CRF model for the extraction and also the subsequent normalization (Fig.~\ref{fig:length-norm}). The performance difference reaches up to 20 points in precision for the extraction of multi-token entities consisting of 10 tokens and 10 points for entities longer than at least 11 tokens.

For both model types, we observe that the performance drop correlates with the length of the entities. In general, there are fewer training instances for longer entities, as shorter entities are more frequent than longer ones with a tail of infrequent but long entities
(Fig.~\ref{fig:length-freq}).
This performance gap between short and long entities is even larger for the normalization which ranges from 85 \fscore for single-token entities to 15 \fscore for entities with more than 10 tokens.
However, as more than half of the entities consist of a single token, the impact of longer entities on the overall \fscore score is limited and, thus, the difference of the BiLSTM-CRF and biaffine models regarding the overall precision is 2 points, even though the biaffine model is better suited for the extraction of longer multi-token entities.

\section{Conclusion}
In this paper, we described our system for the  CANTEMIST shared task to extract, normalize and rank ICD codes from Spanish clinical documents. As neither language nor domain experts, we tested neural sequence labeling, as well as parsing approaches for the extraction, string matching and Levenshtein distance for the normalization and frequency for the ranking.
We found that the best model is based on a biaffine classifier that achieves 85.3 \fscore, 76.7 \fscore and 77.0 MAP for the three tracks, respectively. 
Future work includes the optimization of the extraction models for long multi-token entities. 

\bibliography{references}

\end{document}